\documentclass[journal]{IEEEtran}
\ifdefined\pdfminorversion
\fi
\ifCLASSINFOpdf
\else
\fi
\usepackage[cmex10]{amsmath}
\usepackage{bbm}
\usepackage{optidef}
\usepackage{amssymb,amsfonts}
\usepackage{textcomp}
\usepackage{xcolor}
\usepackage{amsthm}
\usepackage{graphicx}
\usepackage{stfloats}
\usepackage{multirow}
\usepackage{algpseudocode}
\usepackage{svg}
\usepackage[hidelinks]{hyperref}
\hypersetup{
  pdftitle={Learning Agent-based Model Predictive Control for Holistic Vehicle Performance},
  pdfauthor={Jiaming Zhong; Reza Valiollahi Mehrizi; Mohammad Pirani; Chao Yu; Alireza Kasaiezadeh; Yash Vardhan Pant; Amir Khajepour},
  pdfsubject={Author accepted manuscript; IEEE Transactions on Intelligent Transportation Systems; DOI: 10.1109/TITS.2024.3435551}
}

\makeatletter
\newcommand*{\rom}[1]{\expandafter\@slowromancap\romannumeral #1@}
\makeatother

\begin{document}
%
\title{Learning Agent-based Model Predictive Control for Holistic Vehicle Performance}

\author{Jiaming~Zhong,
        Reza~Valiollahi~Mehrizi,
        Mohammad~Pirani,~\IEEEmembership{Member,~IEEE,}
        Chao~Yu,
        Alireza~Kasaiezadeh,
        Yash~Vardhan~Pant,
        and~Amir~Khajepour,~\IEEEmembership{Senior Member,~IEEE}%
\thanks{J. Zhong, R. V. Mehrizi, C. Yu, and A. Khajepour are with the Mechatronic Vehicle Systems (MVS) Lab, Department of Mechanical and Mechatronics Engineering, University of Waterloo, 200 University Ave West, Waterloo ON, N2L3G1 Canada.}%
\thanks{M. Pirani is with the Department of Mechanical Engineering, University of Ottawa, 75 Laurier Ave East, Ottawa, ON, K1N 6N5 Canada.}%
\thanks{A. Kasaiezadeh is with General Motors Company, Warren, MI 48090 USA.}%
\thanks{Y. V. Pant is with the  Control, Learning and Logic (CL2) Lab, Department of Electrical and Computer Engineering, University of Waterloo, 200 University Ave West, Waterloo ON, N2L3G1 Canada.}}

\markboth{AUTHOR ACCEPTED MANUSCRIPT}%
{\MakeLowercase{\textit{Zhong et al.}}: Learning Agent-based Model Predictive Control for Holistic Vehicle Performance}

\newsavebox{\AAMnoticebox}
\sbox{\AAMnoticebox}{\parbox[b]{\textwidth}{\footnotesize\raggedright
\textbf{Author accepted manuscript.} This article was published in
\textit{IEEE Transactions on Intelligent Transportation Systems}, vol.~25, no.~11,
pp.~17482--17492, November~2024. The version of record is available at
\url{https://doi.org/10.1109/TITS.2024.3435551}.\\[0.5ex]
\copyright~2024 IEEE. Personal use of this material is permitted. Permission from IEEE must be obtained for all other uses, in any current or future media, including reprinting/republishing this material for advertising or promotional purposes, creating new collective works, for resale or redistribution to servers or lists, or reuse of any copyrighted component of this work in other works.
}}
\makeatletter
\renewcommand{\@IEEEpubidpullup}{\dimexpr\ht\AAMnoticebox+\dp\AAMnoticebox+\baselineskip\relax}
\makeatother
\IEEEpubid{\usebox{\AAMnoticebox}}

\maketitle

\begin{abstract}
Agent-based model predictive control (AMPC) has recently been proposed as a distributed scheme that collaborates with all agents to achieve optimal holistic performance. However, its optimality highly depends on the prediction accuracy that requires all agents or their contributions to be known, which is too idealistic for actual implementation. This research proposes a novel practical hybrid control scheme - learning agent-based MPC (LAMPC), combining the model-based AMPC approach and data-based learning methods to improve the holistic vehicle performance for multi-agent systems. The Gaussian process regression (GPR) enhanced by an online data management strategy serves as the learning core to predict unknown contributions. A novel multi-step prediction mechanism leverages the GPR learning potential along the horizon. The predicted mean, representing the learned unknown contributions, completes the system model in the MPC for more accurate control. Meanwhile, a stochastic framework is formulated to guarantee control safety and feasibility using soft chance constraints based on the prediction variance. Both simulations and experiments show that,  with the learning capability, LAMPC outperforms the traditional AMPC. LAMPC can achieve higher tracking performance in well-learned scenarios and always guarantee constraint satisfaction even in less-learned scenarios. Moreover, the proposed hybrid control scheme is efficient for real-time implementation and is flexible to any control agent topology.
\end{abstract}


%
\IEEEpeerreviewmaketitle

\section{Introduction}

\subsection{Motivation}

\IEEEPARstart{H}{olistic} vehicle control (HVC) is a method for controlling a vehicle using any single or multiple controllers that can affect vehicle dynamics, such as differential braking, active front steering, torque vectoring, etc. In recent studies, Agent-based MPC (AMPC) has shown high flexibility in adapting to various configurations and robustness against units’ failure in HVC \cite{tang2021agent}. The main idea was that each control module is individually modelled as an interactive agent assigned with a local control task. Agents could be connected in a flexible plug-and-play fashion while emulating centralized MPC performance.

However, the most significant limitation of AMPC is the requirement for all agents to be known, which is often too idealistic in practice. For example, when a “black-box” controller is developed by a third-party supplier, its algorithm and contribution are unknown if there is no specified interface. This lack of information will cause a significant prediction error in vehicle behavior, leading to unexpected control results.

Thanks to increased computational capabilities in modern control systems, data-driven learning-based control techniques have gradually become possible. Notably, the learning-based MPC (LBMPC) has become one of the main trends in modern controller design \cite{zhang2022survey}. This research is motivated to propose a hybrid scheme combining the data-based machine learning approaches with AMPC to solve the problem of any unknown agents existing in a multi-agent system. The learning module predicts unknown information from data, while the MPC provides an optimal solution within safety boundaries.

\IEEEpubidadjcol

\subsection{Literature Review}

\textbf{Learning-based MPC:} In literature, various machine learning methods could be used in control systems, for example, neural networks \cite{adhau2019embedded} and reinforcement learning \cite{lee2018learning}. However, the Gaussian process (GP) is the most popular learning tool in literature. Most of the existing learning patterns in GP-MPCs focus on specific tasks for fixed scenarios specific tasks. For instance, \cite{ostafew2016learning} proposed a LBMPC controller for an autonomous mobile robot to reduce path-tracking errors over repeated traverses along a reference path. 
A robust constraint learning-based nonlinear MPC (RC-LB-NMPC) was proposed in \cite{ostafew2016robust} to track the path of off-road terrain. 
Similar applications in fixed racing scenarios could also be found in \cite{rosolia2017learning} and \cite{hewing2019cautious}. 
However, it must be more practical when applied to diverse driving scenarios. 
Besides, the massive amount of driving data could be a computational disaster regarding the time complexity of GP. Existing sparse approximations \cite{snelson2005sparse}, \cite{titsias2009variational} still cannot meet the real-time requirements in HVC.

\textbf{Robust and stochastic MPC:} The primary limitation of MPC lies in its substantial reliance on the precision of the predictive model. Explicit MPC \cite{borrelli2017predictive}, for example, can reduce the computational burden for control applications, especially for linear time-invariant (LTI) systems and piecewise linear systems.
However, model uncertainties could be naturally introduced if data-driven methods were deployed to describe the system behavior. There are usually two ways to solve the MPC problem with uncertainties. The first is the robust MPC \cite{bemporad2007robust}, which does not model the uncertainties but considers the worst case to guarantee the constraints satisfaction, resulting in conservative control actions. For instance, a minimum robust invariant set of uncertainties should be determined to shrink the constraint in tube robust MPC \cite{riverso2012tube}. Due to the extensive computational cost, only a few robust MPC approaches have been applied in real-time implementation. The other approach is the stochastic MPC \cite{calafiore2012robust}, where the uncertainties are modelled as random elements and a violation of the constraints is accepted below a given probability threshold. Those probabilistic constraints are often called “chance constraints” \cite{campi2019scenario}. This approach is often more advantageous in that it returns less-conservative results. Some stochastic MPCs have already been applied in vehicle control \cite{bichi2010stochastic}, assisted driving \cite{schildbach2015scenario}, collision avoidance \cite{gharavi2023proactive}, and multi-vehicle network control \cite{yan2006constrained}. As a probabilistic model, GPR is often combined with stochastic MPC \cite{hewing2019cautious}. Vehicle stability is safety-critical. It could be dangerous in less-learned scenarios with unreliable predictions. Although some solutions using the robust or stochastic MPC, as summarized above, have been proposed, balancing the constraint satisfaction and not being too conservative is challenging for HVC. 

In summary, challenges in accuracy, efficiency, safety, and feasibility still exist for applying GP-MPC to HVC in practice.

\subsection{Contributions}

This paper proposes a hybrid control scheme - learning agent-based MPC (LAMPC) for multi-agent systems. The main contributions of this work are summarized as follows:

\begin{itemize}
    \item A complete pipeline for this hybrid scheme was provided. With the learning capability, the proposed LAMPC outperforms the traditional AMPC with higher tracking performance and can be applied to any agent topology.
    \item We proposed a data management strategy for multi-step GPR prediction. It achieves efficient real-time GPR inference with precise predictions along the horizon.
    \item We formulated the GP-MPC with a closed-form stochastic scheme to guarantee control safety. Soft chance constraints are used based on the GPR prediction variance. The feedback assumption in uncertainty propagation and soft constraint improves the optimization feasibility.
\end{itemize}

The paper is structured as follows. Section \ref{Preliminaries} first describes preliminary knowledge. The detailed design of the LAMPC scheme is given in Section \ref{Controller Design}. Following an example in Section \ref{Example}, simulation and experiment results are demonstrated in Section \ref{Exp and sim results}. Conclusions are summarized in Section \ref{Conclusion}.


\section{Preliminaries} \label{Preliminaries}

\subsection{Agent-based MPC}

The concept of AMPC proposed in \cite{tang2021agent} adopts a two-stage approach: the first stage models vehicle behavior under net forces and moments $U_{CG}$ at the vehicle center of gravity (CG), and the second stage captures the actuator dynamics from control inputs $U_{i}$ to $U_{CG}$. In AMPC, actuators are usually clustered as agents, which partitions the index set for all actuators into $N_a$ agent subsets as $\mathbb{S}_{U}=\{\mathbb{S}_{1} \cup \mathbb{S}_{1} \ldots \cup \mathbb{S}_{N_a} \}$ with $\mathbb{S}_{i} \cap \mathbb{S}_{j} = \emptyset ~(\forall i \neq j)$. All agents are assumed to be fully connected through the control network. Iterations are usually required to reach an agreement if there is more than one controllable agent in the network. For each agent indexed by $k$, the local optimization in AMPC can be written as,

\begin{mini}|s|[0]
    {{U}_{i, i\in \mathbb{S}_{k}}^j}
    {J_k = \sum_{j=1}^{N_p} \Bigl(N_a {\| y^j - y_{des}^j \|}_{\frac{R_X}{\sqrt{N_a}}}^{2} +  \sum_{i \in \mathbb{S}_{k}} {\| U_i^{j-1} \|}_{{R_{U_i}}}^{2} \Bigr)} 
    {\label{eq.2}}{}
    \addConstraint{X_{CG}^{j+1} = A_d X_{CG}^{j} + B_d \sum_{i \in \mathbb{S}_{k}} U_{CGi}^{j}+ W_{k}^{j} + W_{CGd}^{j}}
    \addConstraint{U_{CGi}^{j} = A_{di} X_{i}^{j} + B_{di} U_{i}^{j} + W_{di}^{j}}
    \addConstraint{y^{j} = C X_{CG}^{j}}
    \addConstraint{X_{CG}^{j} \in \chi}
    \addConstraint{U_{i}^{j} \in \Omega_i, i \in \mathbb{S}_{k}}
\end{mini}
where vehicle states are denoted as ${X}_{CG}$. At the $j$-th step in the horizon, observation and its desired value are denoted as $y^j$ and $y_{des}^j$. $A_d$, $B_d$, $C$, and $W_{CGd}$ are discretized system matrices and disturbances term. $R_X$ and $R_{U_i}$ are penalty weights for states and control inputs, respectively. $\chi$ and $\Omega_i$ are constraint polygons for vehicle states at CG and control action for the $i$-th actuator. $W_k^j \triangleq \sum_{k' \neq k}Y_{k'}^j$ denotes the sum of contributions from all other agents that agent $k$ received, where $Y_{k'}^j \triangleq B_d \sum_{i\in \mathbb{S}_{k'}}U_{CGi}^j$ denotes the contribution by agent $k'$ broadcasting to others.

\subsection{GP Regression}

GP regression (GPR) is a popular non-parametric kernel-based machine learning tool in many engineering applications. Let $\mathcal{D}_n = (\boldsymbol{X}, \boldsymbol{y}) = \{ (\boldsymbol{x}_1, \boldsymbol{y}_1), \ldots, (\boldsymbol{x}_n, \boldsymbol{y}_n) \}$ denotes the dateset where $\boldsymbol{x}_i = {(x_{i1}, \ldots, x_{ip})}^T \in \mathbb{R}^{p}$ is the input variables and $\boldsymbol{y}_i = {(y_{i1}, \ldots, y_{iq})}^T \in \mathbb{R}^{q}$ is the output variables. 
Assume $\boldsymbol{y}_{i} = f(\boldsymbol{x}_{i}) + \boldsymbol\epsilon_i$ where $\boldsymbol\epsilon_i$ are additive noises and follow independent, identically distributed Gaussian distributions with zero mean and variances $\boldsymbol\sigma_n^2$. GP governs the joint Gaussian distribution of outputs $\boldsymbol{y}$ and prediction $f_*$. The mean and the covariance function of $f_*$ could be expressed as,
\begin{equation} \label{eq.4}
\begin{split}
    \overline{f}_* = &\boldsymbol{K}(\boldsymbol{X}_*, \boldsymbol{X}){[\boldsymbol{K}(\boldsymbol{X}, \boldsymbol{X}) + \sigma_n^2 I]}^{-1} \boldsymbol{y} \\
    cov(f_*) = &\boldsymbol{K}(\boldsymbol{X}_*, \boldsymbol{X}_*) -\\ 
    &\boldsymbol{K}(\boldsymbol{X}_*, \boldsymbol{X}){[\boldsymbol{K}(\boldsymbol{X}, \boldsymbol{X}) + \sigma_n^2 I]}^{-1} \boldsymbol{K}(\boldsymbol{X}_*, \boldsymbol{X}_*)
\end{split} 
\end{equation}
where $\boldsymbol{X}_*$ denotes the test inputs. $\boldsymbol{K}(\boldsymbol{X}, \boldsymbol{X}')$ denotes the covariance matrix whose elements $k(\boldsymbol{x}, \boldsymbol{x}')$ are kernels of pairs of data points. The basic GPR algorithm in (\ref{eq.4}) has $\mathcal{O}(n^3)$ time complexity and $\mathcal{O}(n^2)$ memory complexity, where the inverse process of the Gramian matrix ${[\boldsymbol{K}(\boldsymbol{X}, \boldsymbol{X}) + \sigma_n^2 I]}^{-1}$ is the most computational expensive part in GPR \cite{belyaev2014exact}.

\subsection{Stochastic MPC}

For a system subject to a specific constraint $h(X, \delta) \leq 0$ where $X \in \mathbb{R}^N$ is the system state and $\delta \in \mathbb{R}^d$ is the uncertain parameter, the chance constraint can be written as,
\begin{equation} \label{eq.5}
     \Pr(h(X, \delta) \leq 0) \geq 1- \varepsilon_c
\end{equation}
where $\varepsilon_c \in [0,1]$ quantifies the admissible probability of constraint violation. If uncertainty is unbounded, allowing some degree of violation is a necessary but insufficient condition for feasibility, which is also acceptable in practice.

\section{Controller Design} \label{Controller Design}

\begin{figure*}[!t]
\centering
\includegraphics[width=7.14in]{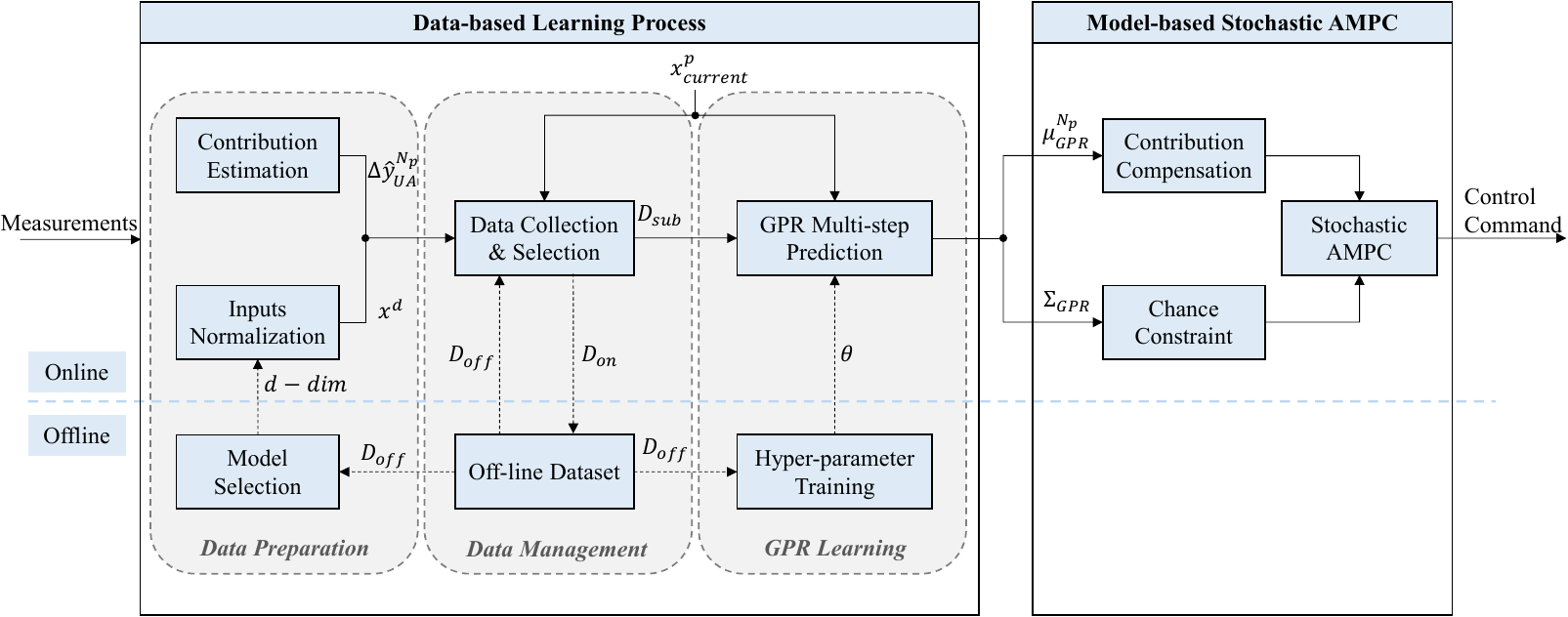}
\caption{The overall structure of the learning agent-based model predictive control (LAMPC). Details are in Section \ref{Controller Design}. 
The $\vartriangle\widehat{y}_{\text{UA}}^{N_p} \in \mathbb{R}^{N_p}$ is the unknown contribution estimated as the output of the data point vector along the horizon from (\ref{eq.7}). The $x^d \in \mathbb{R}^d$ is the normalized data input where $d$ is the selected model variables based on the off-line dataset described in Section \ref{sec_data_preparation}. $D_off$, $D_on$, and $D_sub$ denote the off-line, on-line, and the selected (sub-)datasets, respectively. $\boldsymbol{\theta}$ is the hyper-parameters of the GP model that trained on the off-line dataset. $\mu^{N_p}_{GPR} \in \mathbb{R}^{N_p}$ and $\Sigma_{GPR} \in \mathbb{R}^1$ are the GPR prediction mean values along the horizon and the variance, respectively.
}
\setlength{\tabcolsep}{3pt}
\label{overall_structure}
\end{figure*}

The overall structure of the proposed LAMPC scheme is shown in Fig. \ref{overall_structure}. The learning process consists of data preparation, data management, and GPR learning.
The predicted mean and variance from GPR will finally be used to reconstruct the model-based stochastic AMPC.

\begin{itemize}
    \item Data preparation first pairs the outputs and inputs at the previous step as a new data point for the dataset. Unknown contributions to vehicle dynamics will be estimated as the outputs. Inputs are measurements or vehicle states determined by model selection methods.
    \item Data management includes data collection and subset selection. The new paired data point will be conditionally collected into the dataset. A subset selection is required to accelerate the inference using a small number of data points from the dataset. The dataset could also be saved offline for model selection and hyper-parameter training. 
    \item GPR learning includes offline hyper-parameter training and online inference. The distributions of unknown contributions along the horizon will be predicted with mean values and variance.
    \item The mean and variance of GPR predictions are used separately: The mean is used to complete the system model in MPC as an additive term to make it more accurate, as the total contribution will be compensated. Meanwhile, the soft chance constraints will be formulated based on the propagation of the prediction variance. The re-constructed AMPC will generate optimal and safety control commands for each controllable agent.
\end{itemize}

\subsection{Data Preparation} \label{sec_data_preparation}

\textbf{Estimation of unknown contributions:} The contribution of unknown agents could be estimated from vehicle dynamics, regarding less of the type or inconvenient measurement of the unknown agents. 
The system observation is the combined results from known and unknown agents, which provides a hint for estimating the contribution of the unknown agent as,
\begin{align} \label{eq.7}
\vartriangle\widehat{y}_{\text{UA}}^{(t)} = &\vartriangle{y}_{\text{total}}^{(t)} - \vartriangle{y}_{\text{KA}}^{(t)} \\
= &\ {y}^{(t+1)} - C A_d C^{\dagger} y^{(t)} - C W_{CGd} - C B_d U_{{CG_{\text{KA}}}}^{(t)} \nonumber
\end{align}
where, $C$ is the observation matrix and its pseudo-inverse is denoted as $C^{\dagger}$. The “KA” represents the “known agent” while the “UA” represents the “unknown agent”. $\vartriangle{y}_{\text{KA}}^{(t)} \triangleq C B_d U_{{CG_{\text{KA}}}}^{(t)}$ and $\vartriangle{y}_{\text{UA}}^{(t)} \triangleq C B_d U_{{CG_{\text{UA}}}}^{(t)}$ are the sums of contributions from known and unknown agents, respectively. 
This estimation makes the proposed LAMPC apply to any type and number of unknown agents.

\textbf{Model selection:} Model selection is used to select the most relevant input variables combinations for the dataset. 
Since kernel-based regression, e.g., GPR, could be regarded as a linear smoother \cite{rasmussen2006regression}, the model selection criteria used in multiple linear regression (MLR) also applies to this study. The commonly used criteria can be roughly divided into three categories: 1) predictive risk function, e.g., mean squared prediction error (PMSE); 2) coefficient of determination $R_{adj}^2$; 3) likelihood-based information criteria: e.g., Akaike information criterion (AIC) and Bayesian information criterion (BIC). A dataset including possible input variables needs to be collected first, and then the method described above can be applied offline. The forward-selection could find the smallest subset. It will start from the minimum subset and then add variables while comparing the above criteria for each selection to find the best one. The ranges of the black-box agent’s input variables may vary significantly due to their different physical meanings and units. Therefore, data normalization is needed as a pre-process.

\subsection{Data Management}

\textbf{Data collection:} The online process of data management, mainly the data collection and selection modules, ensures the continuous learning ability and high efficiency of LAMPC. New data will be generated every moment during the controller's real-time operation. Therefore, in case of processing a large amount of repetitive and worthless data, the data collection module should manage the data density while preventing the data from being too aggregated or clustered. The minimum Euclidian distance is used in this study to determine whether a new data point should be added or replaced with an original point. Three strategies shown in Fig. \ref{data_management} (a)-(c):

\begin{figure}
\centering
\includegraphics[width=3.5in]{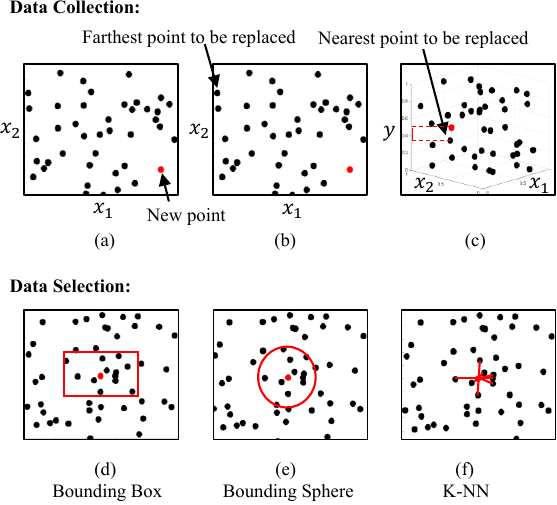}
\caption{Data management strategies. (a)-(c) are data collection strategies: (a) Add the new point; (b) Replace the farthest point with the new point; (C) Replace the nearest point with the new point. (d)-(e) are subset selection strategies: (d) Select points in the bounding box; (e) Select points in the bounding sphere; (f) Select the nearest K neighbor points.}
\setlength{\tabcolsep}{3pt}
\label{data_management}
\end{figure}

\begin{enumerate}
    \item \textit{Add the new input point}: the new point is far from the existing points in the dataset and the dataset is not full. A 2-D example is shown in Fig. \ref{data_management} (a).
    \item \textit{Replace the farthest point}: the new point is far away from the existing points in the training dataset and the dataset is full. A 2-D example is shown in Fig. \ref{data_management} (b). 
    \item \textit{Replace the nearest point}: the new point is very close to it and their output error is significant, which means the new point brings new information that needs updating. A 2-D example is shown in Fig. \ref{data_management} (c). 
\end{enumerate}

\textbf{Subset selection:} As introduced, the inverse process of the Gramian matrix is the most computationally expensive part of GPR. This study adopts three methods for the subset of data (SoD) selection: bounding box, bounding sphere, and K-nearest neighbors (KNN). The core idea is to make the data sparse to reduce the dimension of the Gramian matrix. The schematic diagrams are shown in Fig. \ref{data_management} (d)-(f). All three methods are based on Euclidian distance. The time complexity will reduce from $\mathcal{O}(n^3)$ to $\mathcal{O}(m^3)$ if only $m\leq n$ data points are selected. 

\subsection{GPR Learning}

\textbf{Hyper-parameter training:} As a Bayesian paradigm, the marginal likelihood function is usually to be used to train the hyper-parameters $\boldsymbol\theta$ in GP because it could automatically incorporate a trade-off between model fit and model complexity \cite{rasmussen2005model}. 
The partial derivatives w.r.t. the hyper-parameters $\boldsymbol{\theta}$ are used for solving the optimization problems based on the gradient search method. The training process is processed offline based on an offline dataset. 

\textbf{Multi-step GPR prediction:} In MPC, it is not just one step forward that needs to be predicted but the entire horizon. Conventional GP-MPCs use single-step GPR prediction with some assumptions to predict all the steps along the horizon, but those methods fail to take full advantage of data-based machine-learning approaches. Therefore, some studies (e.g., in \cite{suradhaniwar2021time}) suggest a multi-step prediction method where GPR predicts all steps along the horizon without artificial assumptions. However, these traditional multi-step prediction methods are essentially iterations of multiple single-step predictions. Each step's prediction requires the previous step's prediction value, which means the input matrix $\boldsymbol{X}$ needs to be updated at each prediction step along the horizon. Therefore, those methods are feasible but could be a computational disaster for real-time implementation, especially when the horizon is long.

\begin{figure}
\centering
\includegraphics[width=3.5in]{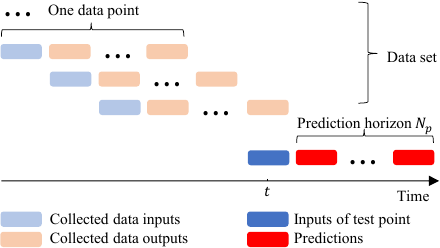}
\caption{Schematic diagram of data collection for multi-step prediction.}
\setlength{\tabcolsep}{3pt}
\label{multi_prd}
\end{figure}

This study proposes a novel efficient multi-step prediction method. The main idea is, at each time step, to predict all steps along the horizon at once using a single input matrix $\boldsymbol{X}$, as shown in Fig. \ref{multi_prd}. The key is the way of collecting the data. For example, at the current time step $t$ with $N_p$-step horizon, the collection of input matrix $\boldsymbol{x}_i$ starts from the time step $t-N_p$ and goes earlier. At the same time, the output data is a vector $\boldsymbol{y}_i$ with $N_p$ elements. In this case, the input matrix $\boldsymbol{X}$ is fixed for predicting all the steps along the horizon. Thus, the inverse process of the Gram matrix only needs once at each time step. Besides, the GPR mean prediction can be calculated by simple matrix operations, where replace the vector $\boldsymbol{y}$ ($m\times1$) with the matrix $\boldsymbol{Y}$ ($m\times N_p$) in (\ref{eq.4}).

\subsection{Soft Stochastic MPC Scheme}

\textbf{Contribution compensation:} 
For each known controllable agent k, the system equation at CG can be completed using the GPR prediction results as,
\begin{equation} \label{eq.10}
\begin{aligned}
    X_{CG}^{j+1} = &A_d X_{CG}^{j} + B_d \sum_{i \in \mathbb{S}_{k}} U_{CGi}^{j} + W_{k}^{j} \\
    &+ W_{CGd}^{j} + X_{CG,\text{UA},prd}^{j}
\end{aligned}
\end{equation}
where $X_{CG,\text{UA},prd}^{j} \sim \mathcal{N}(\mu_{GPR}^j, \Sigma_{GPR})$ denotes the predicted unknown contribution.
$\mu_{GPR}^j = \overline{f}_*^j$ is the predicted mean value and $\Sigma_{GPR} = cov({f}_*)$ is the predicted variance in (\ref{eq.4}).

\textbf{Uncertainty propagation:} Since the $X_{CG,\text{UA},prd}^{j}$ acts as an additional disturbance, the uncertainty from GPR predictions will propagate along the horizon through the system equation in (\ref{eq.10}). It is known that the uncertainty will increase along the horizon if the open-loop propagation is used, which is a risk of infeasibility if the uncertainty from GPR is significant (e.g., driving into a new scenario without previous experience.) or the horizon $N_p$ is long. Therefore, the closed-loop propagation with a linear feedback assumption is considered, which can be written for the $i$-th actuator in agent $k$ as,
\begin{equation} \label{eq.12}
\renewcommand{\arraystretch}{1.3}
\begin{array}{c}
    U_i^j = v_i^j + K_i(X_{CG}^j - z_{CG}^j) = v_i^j + K_i e_{CG}^j, i\in \mathbb{S}_{k}\\
    e_{CG}^{j+1} = \Bigl(A_d + B_d\sum_{i\in \mathbb{S}_{k}} B_{di} K_i\Bigr) e_{CG}^j + \omega
\end{array}
\end{equation}
where, $v_i^j$ is the nominal control action, $z_{CG}^j$ is the nominal vehicle states, $e_{CG}^j\sim \mathcal{N}(0,\Sigma_e^j)$ and $\omega\sim\mathcal{N}(0,\Sigma_{GPR})$ are unbiased disturbance terms with the uncertainty of error dynamics and GPR, respectively. $K_i$ is the feedback matrix. Thus, the system dynamics on vehicle states $X_{CG} = z_{CG} + e_{CG}$ could be regarded as the combination of the noise-free nominal dynamics of $z_{CG}$ and the error dynamics $e_{CG}$. By choosing feedback $K$ to stabilize dynamics $\Tilde{A}_d = A_d + B_d \sum_{i\in \mathbb{S}_{k}} (B_{di}K_i)$, the increase of $e_{CG}$ along the horizon will asymptotically close to zero. 

The unbounded distribution $\omega$ from GPR can be turned into a bounded polygon $\xi_{GPR}(\varepsilon_c)$ if manually set a confidence level $(1-\varepsilon_c)$ in the quantile function, where $\varepsilon_c$ is the admissible probability of constraint violation (e.g., $2.5\%$) in the chance constraint in (\ref{eq.5}). The uncertainty propagated from GPR to the error dynamics $e_{CG}^{j} \in \mathcal{E}_{CG}^{j}$ could be derived as,

\begin{equation} \label{eq.13}
    \mathcal{E}_{CG}^{j} = \oplus_{j=1}^{N_p} (\Tilde{A}_d^j \xi_{GPR}(\varepsilon_c))
\end{equation}
where $\oplus$ denotes the Minkowski sum. The uncertainty of the error dynamics will be further propagated to vehicle states and control actions. Thus, the constraints should be shrunken as,
\begin{equation} \label{eq.14}
\renewcommand{\arraystretch}{1.3}
\begin{array}{c}
    z_{CG}^j \in \chi \ominus \mathcal{E}_{CG}^{j} \\
    v_i^j \in \Omega_i \ominus K_i \mathcal{E}_{CG}^{j}
\end{array}
\end{equation}
where $\ominus$ denotes the Minkowski subtraction. 

\textbf{Soft stochastic LAMPC:} The stochastic LAMPC is now an optimization on the noise-free nominal system of $v_i^j$ and $z_{CG}^j$, with shrunken constrain polygons at each time step. The optimal control law for the $i$-th actuator in agent $k$ should be,
\begin{equation} \label{eq.15}
    U_i^{j*} = v_i^{j*} + K_i (X_{CG}^j - z_{CG}^{j*})
\end{equation}
which, in fact, is intractable to solve because the states $X_{CG}^j$ in unknown in the horizon. However, in MPC context, only the first step needs to be applied as the control input, where $U_i^{0*} = v_i^{0*}$ because $X_{CG}^0 = z_{CG}^{0*}$. Thus, we can still use the batch formulation to solve the optimization in real time and only apply the $v_i^{0*}$ as the control action to the system.

Although the feedback assumption is applied, the feasibility of the optimization is still not guaranteed. Therefore, softened chance constraints (SCC) are further considered to guarantee the optimization feasibility. Defining $\mathcal{E}_z^j$ and $\mathcal{E}_{v,i}^j$ are the slack polygons for $z_{CG}^j$ and $v_i^j$ respectively, the LAMPC with soften chance constraints (SCC) can be written as,
\begin{mini}|s|[0]
    {\begin{array}{c}
    v_i^{j-1}\\
    z_{CG}^j
    \end{array}}
    {J_k = \sum_{j=1}^{N_p} 
    \begin{pmatrix*}[l]
    N_a {\| y^j - y_{des}^j \|}_{\frac{R_X}{\sqrt{N_a}}}^{2} + {\| \varepsilon_z^j \|}_{{R_z}}^{2}\\
    +\sum\limits_{i \in \mathbb{S}_{k}} { {\| v_i^{j-1} \|}_{{R_{U_i}}}^{2} + {\| \varepsilon_{v,i}^j \|}_{{R_{v,i}}}^{2} }
    \end{pmatrix*}
    }     
    {\label{eq.16}}{}
    \addConstraint{\begin{matrix*}[r]
    z_{CG}^{j+1} = A_d z_{CG}^{j} + B_d \sum_{i \in \mathbb{S}_{k}} U_{CGi}^{j} + W_{k}^{j} \\
    + W_{CGd}^{j} + \mu_{GPR}^j
    \end{matrix*}}
    \addConstraint{U_{CGi}^{j} = A_{di} X_{i}^{j} + B_{di} v_{i}^{j} + W_{di}^{j}}
    \addConstraint{y^{j} = C z_{CG}^{j}}
    \addConstraint{z_{CG}^0 = X_{CG}^0, \text{given} }
    \addConstraint{z_{CG}^{j} \in \chi \ominus \mathcal{E}_{CG}^j \oplus \mathcal{E}_z^j}
    \addConstraint{v_{i}^{j} \in \Omega_i \ominus K_i\mathcal{E}_{CG}^j \oplus \mathcal{E}_{v,i}^j}
    \addConstraint{0 \leq \varepsilon_z^j}
    \addConstraint{0 \leq \varepsilon_{v,i}^j}
    \addConstraint{i\in \mathbb{S}_{k}, j=1, \ldots, N_p}
\end{mini}
where $R_z$ and $R_{v,i}$ are heavy penalties on slack polygons. The LAMPC in (\ref{eq.16}) will always be feasible and as safe as possible based on the probabilistic prediction results.

\textbf{Discussion of stability:} Although it has been intensively studied recently, the stability of data-driven control systems is still an ongoing research field. As an uncertain system, the stability of a learning-based MPC usually consists of nominal stability and robust stability \cite{maiworm2018stability}. The former is defined as the asymptotic stability for the nominal control system without uncertainty, while the latter considers the bounded effect of the uncertainty on the system. 
For the nominal stability of a finite-horizon MPC optimization problem, the standard approach is to define a proper terminal cost as a control Lyapunov function (CLF) with a
terminal constraint \cite{maiworm2018stability}\cite{mayne2001control}. 
For robust stability, some recent works \cite{maiworm2021online}\cite{rose2023learning} have proposed several learning-based controllers with data selection methods to achieve robust stability. However, these approaches are based on the assumption of bounded uncertainty, which is unsuitable for the GP-MPC. Moreover, they usually adopt iterative data management methods, which is another challenge for real-time implementations.
We will extend this work and continue to study hybrid data/model-based control methods that can ensure stability in the future.

\section{Example: Vehicle Stability Control} \label{Example}

\subsection{Agent Configuration}

System matrices in (\ref{eq.2}) for vehicle stability control are,
\begin{equation} \label{eq.17}
\renewcommand{\arraystretch}{1.3}
\begin{array}{c}
X_{CG}=\begin{bmatrix} v\\r \end{bmatrix},
U_{CG}=\begin{bmatrix} F_{yCG}\\M_{zCG} \end{bmatrix},
W_{CG}=\begin{bmatrix} 0\\0 \end{bmatrix},
\\
A=\begin{bmatrix*}[c] 0&-u\\0&0 \end{bmatrix*},
B=\begin{bmatrix*}[c] 1/m&-u\\0&1/I_z \end{bmatrix*},
C=\begin{bmatrix} 0&1 \end{bmatrix}
\end{array}
\end{equation}
where, $u$ and $v$ denote the longitudinal and lateral speed, respectively. $r$ denotes the yaw rate. $F_{yCG}$ and $M_{zCG}$ denote the net forces and moments at the vehicle CG, respectively. $m$ and $I_z$ denote the curb weight and yaw inertia, respectively. 

Four agents are considered in this example as in Table \ref{table.1}. The Rear differential torque (RDT) is the controllable agent with the designed controller. The driver steering (STR), and the driver torque (DT) agent are the white-box agents, which are uncontrollable, but control efforts are known. The front differential torque (FDT) agent is a black-box agent with unknown contributions to vehicle behaviors.

\begin{table}[!ht]
\renewcommand{\arraystretch}{1.3}
\caption{Agent Configuration}
\label{table.1}
\centering
\begin{tabular}{c|c|c}
\hline\hline
Type & Agent & $U_i$ \\
\hline
Controllable & Rear differential torque (RDT) & $Q_{RDT, \{ rl,rr \}}$ \\
\hline
White-box & Driver steering (STR) & $F_{y, \{ fl,fr,rl,rr \}}$ \\
\hline
White-box & Driver torque (DT) & $Q_{DT, \{ fl,fr,rl,rr \}}$ \\
\hline
Black-box & Front differential torque (FDT) & $Q_{FDT, \{ fl,fr \}}$ \\
\hline\hline
\end{tabular}
\end{table}

\textbf{Known agents:} Agent RDT is the controllable agent, and its contributions in (\ref{eq.2}) should be represented as ($k \triangleq RDT$),
\begin{equation} \label{eq.18}
\renewcommand{\arraystretch}{1.3}
\begin{array}{c}
    Y_{RDT} = B_d U_{CGRDT} \\
    W_{RDT} = B_d U_{CGSTR} + B_d U_{CGDT}
\end{array}
\end{equation}
where $U_{CGRDT}$, $U_{CGSTR}$, and $U_{CGDT}$ denote the contribution from RDT, STR, and DT agents, respectively. Mathematical models of these agents are,
\begin{equation} \label{eq.19}
\begin{aligned}
    &U_{CGRDT} = \sum\limits_{i\in\{rl,rr\}} {\left[\frac{1}{R_{ei}}, \frac{T_{wi}}{R_{ei}}\right]}^T Q_{RDTi} \\
    &\begin{array}{r}
    U_{CGDT} = \sum\limits_{i\in\{fl,fr\}} {\left[\frac{\sin{\delta_i}}{R_{ei}}, \frac{T_{wi}\cos{\delta_i} + a_i\sin{\delta_i}}{R_{ei}}\right]}^T Q_{DTi} \\
    + \sum\limits_{i\in\{rl,rr\}} {\left[\frac{1}{R_{ei}}, \frac{T_{wi}}{R_{ei}}\right]}^T Q_{DTi}
    \end{array}\\
    &U_{CGSTR} = \sum\limits_{i\in\{fl,fr\}} {\left[1, a_i\right]}^T F_{yi},
    a_i \triangleq \begin{cases} +a, \text{front} \\ -b, \text{rear} \end{cases}
\end{aligned}
\end{equation}
where $Q_{RDT}$ and $Q_{DT}$ are the torques from RDT and DT agents, respectively. $F_y$ is the lateral tire force from the estimation module. $a$ and $b$ are the distances from the front and rear axle to CG, respectively. $T_w$ is the wheelbase. $R_e$ is the wheel radius. $\delta$ is the wheel steering angle. $i\in\{fl,fr,rl,rr\}$ represent the front-left, front-right, rear-left, and rear-right wheels, respectively.

\textbf{Unknown agent:} As a ground truth for method verification, the model of the FDT agent is manually set in simulations and experiments ($Q_{{FDT}_{fr}} = -Q_{{FDT}_{fl}}$). A commonly used control architecture, namely a feedback control plus a feed-forward (FF) control, is adopted in this study as the control model of FDT. The feedback control is realized by a PI controller, whose input is the error between the measured and desired yaw rate ($e_r=r_{des}-r_{mes}$). The feed-forward control is proportional to the desired yaw angular acceleration $\dot{r}_{des}$:
\begin{equation} \label{eq.20}
\begin{aligned}
    Q_{{FDT}_{fr}} = K_P e_r + K_I \int_0^t e_r(\tau)d\tau + K_{FF} \cdot \dot{r}_{des}
\end{aligned}
\end{equation}
where $K_p$, $K_I$, $K_{FF}$ are the proportional, integral, and feed-forward gains, respectively. The ground truth contribution from the black-box FDT agent $U_{CGFDT}$ can be summarized as,
\begin{equation} \label{eq.23}
    U_{CGFDT} = f(r_{mes}, U_{CGFDT}^{prev}, r_{des}, r_{des}^{prev})
\end{equation}
where $U_{CGFDT}^{prev}$ and $r_{des}^{prev}$ are the contribution and the desired yaw rate at the last time step, respectively. 

\subsection{LAMPC Controller}

\textbf{Vehicle stability control:} Vehicle stability control is usually formulated to track the desired yaw rate with the limitation of the maximum side slip angle at the rear axle and the maximum yaw rate. The desired yaw rate in this example is calculated using the linear bicycle model:
\begin{equation} \label{eq.21}
\renewcommand{\arraystretch}{1.3}
\begin{aligned}
    X_{CG}^{des} = \begin{bmatrix} v_{des} \\ r_{des} \end{bmatrix}
    = \begin{bmatrix} 0 \\ \text{sign}(\delta_f) \min \left( |\frac{u}{L_b+K_{us}u^2} \delta_f |, \frac{\mu g}{u} \right) \end{bmatrix}
\end{aligned}
\end{equation}
where $L_b$ is the wheelbase of the vehicle. $K_{us}$ is the under-steering coefficient. $\mu$ is the road friction from the estimation module. $\delta_f$ is the equivalent front wheel turning angle. $g$ is the gravity acceleration. The constraints, which include the actuator limitation of RDT and stability requirements, can be written as,

\begin{equation} \label{eq.22}
\renewcommand{\arraystretch}{1.3}
\begin{array}{c}
\Omega \triangleq: \begin{bmatrix} I\\-I \end{bmatrix} Q_{RDT} \leq L\\
L = \begin{bmatrix}
    Q_{max}-\max\{Q_{DT,fr}, Q_{DT,fl}, 0 \} \\
    \min\{Q_{DT,fr}, Q_{DT,fl}, 0 \} - Q_{min}
\end{bmatrix}\\
\chi \triangleq: F=\begin{bmatrix}
    1/u & -b/u \\ -1/u & b/u \\ 0 & 1 \\ 0 & -1
\end{bmatrix} \leq 
G=\begin{bmatrix}
    \alpha_{rear,max} \\ \alpha_{rear,max} \\ r_{max} \\ r_{max}
\end{bmatrix}
\end{array}
\end{equation}
where $Q_{max}$ and $Q_{min}$ are the maximum and minimum capacity of the wheel motor. $\alpha_{rear,max} \geq 0$ and $r_{max} \geq 0$ are the maximum allowable side slip angle at the rear axle and yaw rate, respectively.

\textbf{Model selection and training:} 
The variable combination with the best statistical learning criteria is selected as the input variable for data collection. 
As we tested using an offline dataset, the ground-truth input set of the black-box FDT agent will always have the lowest PMSE, highest $R_{adj}^2$, and lowest AIC values. The model selection method used in this study can accurately identify the correct input variables in (\ref{eq.23}).

The squared exponential (SE) kernel is used in this example:
\begin{equation} \label{eq.24}
    k(\boldsymbol{x}, \boldsymbol{x}') = \sigma_f^2 \exp{\left[ -\frac{1}{2} \sum_{i=1}^d \frac{{(x - x')}^2}{l_i^2} \right]}
\end{equation}
where $\sigma_f$ denotes the signal standard deviation (SD) and $l_i$ denotes the length scales of each input variable. Based on the results of model selection, $d=4$. Thus, there are six hyper-parameters in total that need to be trained, including the noise standard deviation $\sigma_n$ in (\ref{eq.4}).

\textbf{Soft stochastic LAMPC:} The uncertainty from GPR prediction will be propagated to each prediction step through (\ref{eq.13}). Following a common approach that the state, control input, and disturbance are approximated as jointly Gaussian distributed at each step, affine transformations of the Gaussian distribution \cite{hewing2019cautious} are used for the propagation in (\ref{eq.14}) from error dynamics to system states and control actions. By setting an admissible probability of constraint violation $\varepsilon_c$, the shrunken biases at prediction step $j$ on constraints are,
\begin{equation} \label{eq.25}
\begin{aligned}
    \vartriangle G^j =& z_p \sqrt{F\Sigma_e^j F^T} \\
    \vartriangle Q_i^j =& z_p \sqrt{K_i \Sigma_e^j K_i^T}, i=RDT
\end{aligned}
\end{equation}
where $\Sigma_e^j$ is the variance of error dynamics in (\ref{eq.13}) at step $j$. $z_p$ is the quantile of Gaussian distribution where $p=1-\varepsilon_c$. 

The shrunken constraints are further relaxed as soft chance constraints to guarantee the feasibility: 
\begin{equation} \label{eq.26}
\begin{aligned}
\Omega^j \triangleq: \begin{bmatrix} I\\-I \end{bmatrix} v_{RDT}^{j-1} &\leq L - \begin{bmatrix} \vartriangle Q_{RDT}^j - \varepsilon_{v,RDT}^j\\\vartriangle Q_{RDT}^j - \varepsilon_{v,RDT}^j \end{bmatrix} \leq L \\
\chi^j \triangleq: F z_{CG}^j &\leq G^j - (\vartriangle G^j - \varepsilon_z^j)
\end{aligned}
\end{equation}
where $z_{CG}^j$ and $v_{RDT}^{j-1}$ are noise-free nominal vehicle state and control inputs. $\varepsilon_z^j \geq 0$ and $\varepsilon_{v,RDT}^j \geq 0$ are slack variables on state constraints and control constraints, respectively, corresponding to the slack polygons $\mathcal{E}_z^j$ and $\mathcal{E}_{v,RDT}^j$.

\section{Real-time Implementation and Results} \label{Exp and sim results}

Based on the above example, simulation and experimental studies were conducted to demonstrate the features of the proposed LAMPC control scheme. 
The controller is developed in Matlab/Simulink with parameters shown in Table \ref{table.3}.
A high-fidelity calibrated Carsim model of this vehicle from a previous experimental study \cite{hashemi2020vehicle} is adopted for simulations, with the controller running at a frequency of 20Hz.
Real-time experiments were conducted on an electric vehicle with four independent-drive electric motors, with the controller running in real-time at a frequency of 50Hz on a laptop with an i7-12700H CPU and 16 GB RAM. The controller ran in Matlab Real-time Desktop model and sent the control command to motor drivers through the CAN bus at every time step.

\begin{table}[!ht]
\renewcommand{\arraystretch}{1.3}
\caption{Platform and Controller Settings}
\label{table.3}
\centering
\begin{tabular}{c|c|c}
\hline\hline
Symbol & Parameter Description & Value \\
\hline
$N_p$ & Prediction horizon & $10$ \\
$R_X$ & Weight for yaw rate tracking & $5\times{10}^6$ \\
$R_{U_{RDT}}$ & Control weight on RDT agent & $0.3$ \\
$R_z, R_v$ & Slack weight on soft constraints & $1\times{10}^{10}$ \\
$Q_{max}$ & Maximum capacity of the wheel motor & $1760$ Nm \\
$Q_{min}$ & Minimum capacity of the wheel motor & $-1760$ Nm \\
$K_P$ & Proportional gain for FDT agent & $100$ \\
$K_I$ & Integral gain for FDT agent & $0.1$ \\
$K_{FF}$ & Feed-forward gain for FDT agent & $0.5$ \\
$\alpha_{rear,max}$ & Maximum side slip angle at the rear axle & $10$ deg \\
$r_{max}$ & Maximum yaw rate & $50$ deg/s \\
$\varepsilon_c$ & Admissible probability of violation & $2.5 \% $\\
\hline\hline
\end{tabular}
\end{table}

The proposed method was first tested in the “less-learned” scenario, where there was not enough data with similar experience for GPR prediction, to verify the safety using the stochastic chance constraint. Then several tests in the “well-learned” scenario were conducted to demonstrate the improvement of the proposed LAMPC compared to the conventional AMPC.

\subsection{Less-learned Scenario: Safety Guarantee}

\textbf{Simulations:} 
A double lane change (DLC) maneuver was operated at about 65 km/h on a high friction surface. Control agents were configured as in Table \ref{table.1}. 
The dataset of the controllable RDT agent started empty, meaning there is not enough experience for this DLC. Three controllers have been applied for the RDT agent: 
1) AMPC controller; 
2) LAMPC controller with normal constraints (NC), i.e. the stability constraints without considering the learning uncertainties as described in (\ref{eq.22});
3) LAMPC controller with soft chance constraints (SCC) as described in (\ref{eq.26}).

\begin{figure}
\centering
\includegraphics[width=3.49in]{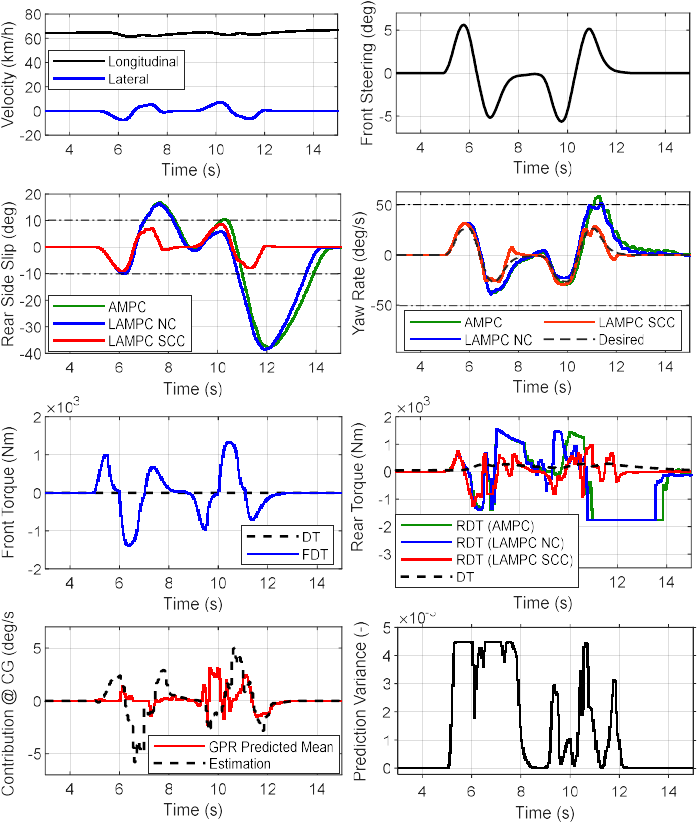}
\caption{Simulation results of a DLC in the less-learned scenario.}
\setlength{\tabcolsep}{3pt}
\label{sim_lls}
\end{figure}

\begin{figure}[!ht]
\centering
\includegraphics[width=3.49in]{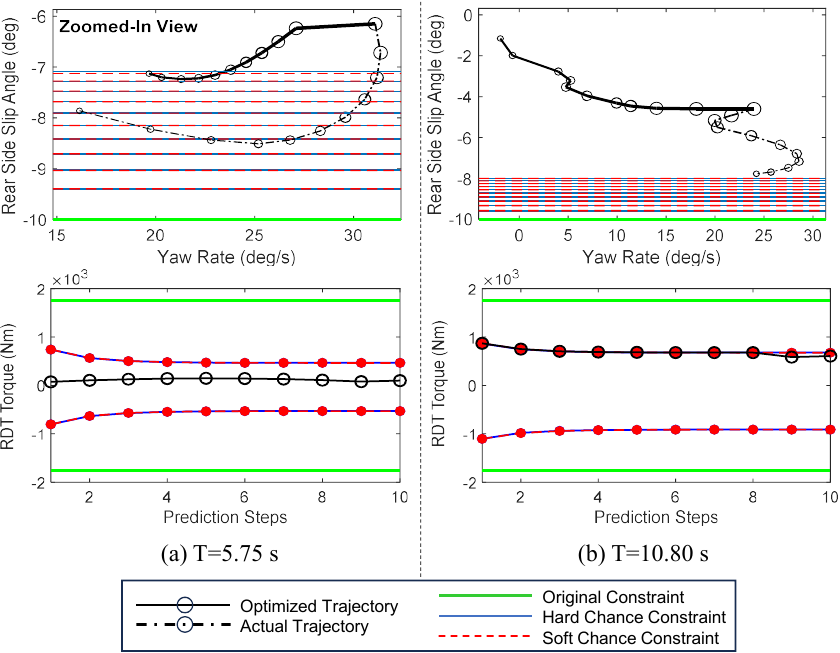}
\caption{Optimization results in the less-learned DLC at T=5.75 and T=10.80.}
\setlength{\tabcolsep}{3pt}
\label{sim_lls_clip}
\end{figure}

The simulation results of the LAMPC with SCC are shown in Fig. \ref{sim_lls}, where the results of AMPC and LAMPC with normal constraints are also compared. 
It is easy to notice that when the RDT agent was controlled by AMPC or LAMPC with NC, the vehicle's rear slip angle and yaw rate exceeded the limits. Remarkably, the absolute value of the maximum side slip angle has reached 40 deg, which is very dangerous in reality. This violation is because the black-box FDT agent significantly impacts vehicle dynamics, but GPR cannot predict its contribution, shown as the significantly predicted error and significant variances. The RDT agent will therefore generate far more torque than is required. However, when the RDT agent was controlled by the proposed LAMPC with SCC, the rear slip angle and the yaw rate were kept within constraints, ensuring the vehicle was stable and safe.

\textbf{Details:} 
Two typical optimization results are shown in Fig. \ref{sim_lls_clip}. 
The prediction uncertainty is propagated as constraints shrink simultaneously on both state and control inputs. 
The original constraints (green solid lines) were first shrunken to hard chance constraints (blue dashed lines) and then relaxed as softened chance constraints (red dashed lines) at each step along the horizon. 
At time T=5.75, the state trajectory was constrained; At time T=10.80, the control input was constrained. The actual state trajectories were all guaranteed not to exceed the original constraint under the $\varepsilon_c$.

\subsection{Well-learned Scenario: Performance Improvement}

\begin{figure}
\centering
\includegraphics[width=3.49in]{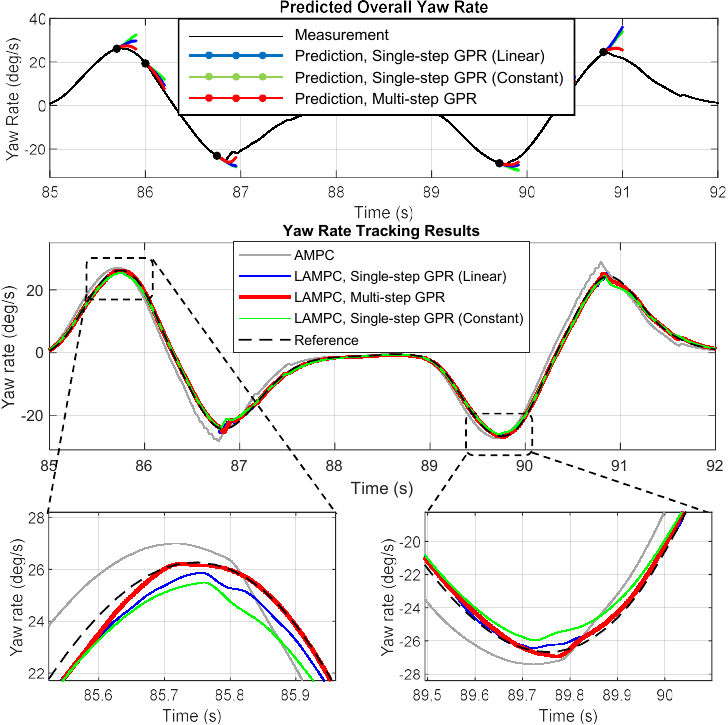}
\caption{Comparison of the yaw rate prediction and tracking performance in the last DLC in the well-learned scenario.}
\setlength{\tabcolsep}{3pt}
\label{sim_wls_prd}
\end{figure}

\textbf{Simulations:} The second scenario investigates the control performance of the proposed LAMPC to see if the learning results are reliable enough. 
A detailed comparison of using different controllers in a well-learned DLC maneuver is shown in Fig. \ref{sim_wls_prd}. The dataset already has enough data collected from similar DLCs. The predictions of the compensated overall yaw rate are first compared. The predictions using the proposed multi-step GPR (red) were significantly closer to the actual measured yaw rate than single-step GPR with assumptions (blue and green), especially when the yaw rate changes fast. With more accurate predictions, the tracking performance of the proposed LAMPC based on multi-step GPR is better than the traditional AMPC and those using the single-step GPR, especially when the yaw rate changes drastically.

Besides, a comparison of the computational time consumption between using the single-step GPR and multi-step GPR is given in TABLE \ref{table.time}, which is the average time consumptions counted by the Simulink built-in toolkit in the simulation shown in Fig. \ref{sim_wls_prd}. As we can see, adopting the proposed multi-step GPR prediction only slightly increases the time consumption compared to single-step GPR prediction, mainly in data management and GPR prediction. However, the increase is not significant and, therefore, will not affect the overall real-time efficiency. In addition, the overall time consumption of all modules can fully meet the real-time requirements.

\begin{table}
\renewcommand{\arraystretch}{1.3}
\caption{Comparison of time consumption of modules}
\label{table.time}
\centering
\begin{tabular}{c|c|c}
\hline\hline
Module & 
{\renewcommand{\arraystretch}{1.1}\begin{tabular}[c]{@{}c@{}}Prediction using the\\single-step GPR\end{tabular}}
& 
{\renewcommand{\arraystretch}{1.1}\begin{tabular}[c]{@{}c@{}}Prediction using the\\proposed multi-step GPR\end{tabular}} \\
\hline
MPC Solving & 0.25 ms & 0.27 ms \\
Data Management & 1.26 ms & 1.44 ms \\
GPR Inference & 0.38 ms & 0.46 ms \\
Other & 1.99 ms & 1.90 ms \\
\hline\hline
\end{tabular}
\end{table}

\begin{figure}
\centering
\includegraphics[width=2.0in]{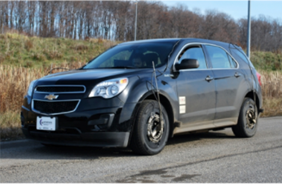}
\caption{Test vehicle: the Chevrolet Equinox electric vehicle.}
\setlength{\tabcolsep}{3pt}
\label{test_vehicle}
\end{figure}

\textbf{Experiments:} Experiments on the real vehicle were also conducted where the LAMPC controller ran at an even higher control frequency (50 Hz). 
The test vehicle in this study is the Chevrolet Equinox electric vehicle as shown in Fig. \ref{test_vehicle}, whose main parameters are specified in Table \ref{table.4}. In addition to driver-controlled drive, braking and steering, this test vehicle also enables independent differential torque control for the front and rear axles as torque can be individually controlled for all four wheels, as well as an additional active front steering input. At the same time, this vehicle also provides a wealth of sensor interfaces, including GPS, IMU, wheel speed, etc., to facilitate the provision of more accurate reference speed estimation and other key signals through sensor fusion.

\begin{table}
\renewcommand{\arraystretch}{1.3}
\caption{Specification of the Test Vehicle}
\label{table.4}
\centering
\begin{tabular}{c|c|c}
\hline\hline
Parameter & Description & Value \\
\hline
$m$ & Vehicle mass & $2272$ kg \\
$I_z$ & Vehicle inertia & $4600~\text{kg}\cdot{\text{m}}^2$ \\
$I_w$ & Vehicle track width & $1.6$ m \\
$a$ & Front distance from axles to CG & $1.42$ m \\
$b$ & Rear distance from axles to CG & $1.43$ m \\
$R_e$ & Effective radius of wheels & $0.351$ m \\
$C_{\alpha}$ & Cornering stiffness of tires & $130000$ N/rad \\
$R_w$ & Steering ratio & $18.5$ \\
\hline\hline
\end{tabular}
\end{table}

\begin{figure*}[b]
\centering
\includegraphics[width=7.13in]{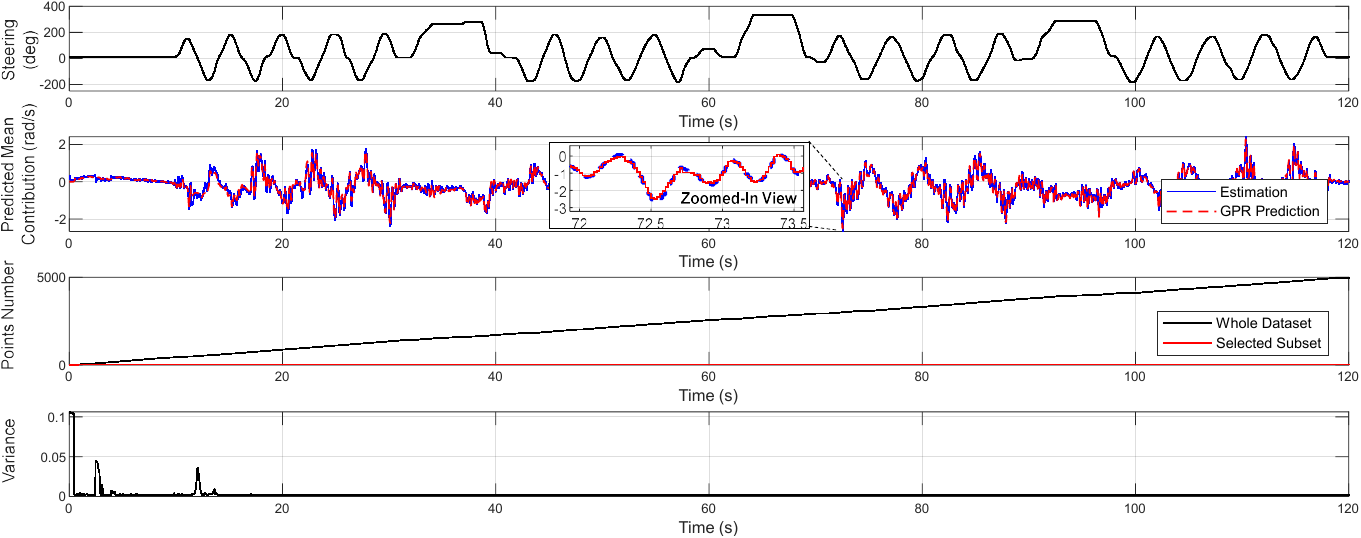}
\caption{Online learning process of the Sinewave experiment.}
\setlength{\tabcolsep}{3pt}
\label{exp_wls_learning}
\end{figure*}

Several consecutive sinewave maneuvers were implemented. The learning process is shown in Fig. \ref{exp_wls_learning}. The proposed learning scheme can accurately predict the contribution from the unknown agents. As the number of collected data kept growing, the shortest and average distance to the test point decreased in the 20-point subset. Therefore, the predictions continually became more accurate, which could be proved by decreasing the prediction variance.

\begin{figure*}
\centering
\includegraphics[width=7.0in]{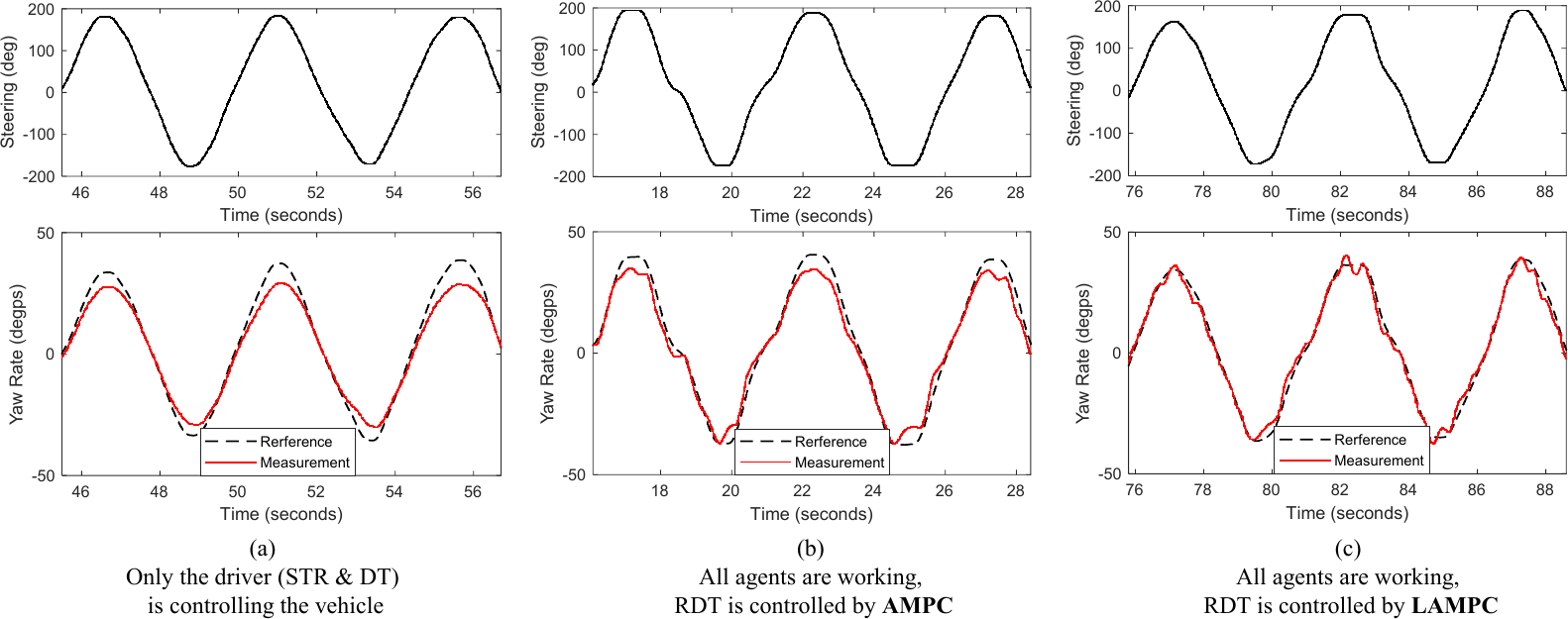}
\caption{Comparison results of yaw rate tracking performance in the Sinewave experiment.}
\setlength{\tabcolsep}{3pt}
\label{exp_wls_compare}
\end{figure*}

Three similar sinewave maneuvers with different settings were also implemented for comparison, as shown in Fig. \ref{exp_wls_compare}: (a) RDT and FDT agents were turned off, and the vehicle was only controlled by the DT and STR agents; (b) All agents are activated where FDT was the black-box agent and only a traditional AMPC controller was working on the RDT agent; (c) All agents are activated where FDT was the black-box agent and the proposed LAMPC with learning function is working on the RDT agent. It is shown that when all agents are activated, and the traditional AMPC controller controls the controllable RDT agent, the tracking performance is improved compared to the first experiment, where only the driver controls the vehicle. However, without the information of the black-box agent, there is still a significant tracking error. In the last experiment, since the proposed LAMPC controller can learn the contribution of the black-box FDT agent from data, the measured yaw rate is always closer to the reference. 

\begin{figure}
\centering
\includegraphics[width=3.49in]{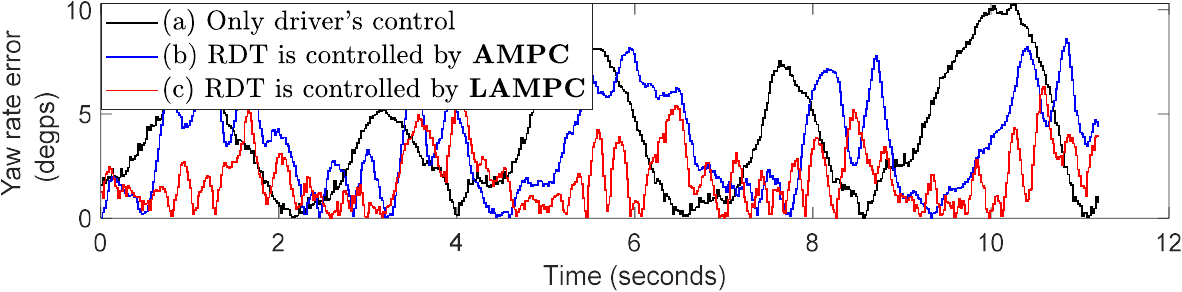}
\caption{Synchronized absolute yaw rate errors during Sinewave experiments.}
\setlength{\tabcolsep}{3pt}
\label{MSE}
\end{figure}

Quantitatively, as shown in Fig. \ref{MSE}, the mean-squared error (MSE) of yaw rates was reduced from 18.2 $\text{deg}/\text{s}^2$ to 5.9 $\text{deg}/\text{s}^2$ when the RDT agent is controlled by the LAMPC.

\section{Conclusion} \label{Conclusion}

This study proposed a general and practical hybrid control scheme: learning agent-based MPC (LAMPC). It has been proven by simulations and experiments that the proposed LAMPC scheme has the following advantages: 

1) \textit{High tracking performance}. The tracking performance was significantly improved in a well-learned scenario based on reliable learning results.

2) \textit{Guaranteed safety}. Stochastic chance constraint is used to guarantee the safety of constraint satisfaction. The feedback assumption in uncertainty propagation and soft constraint help to improve and ensure the optimization feasibility.

3) \textit{Real-time efficiency}. This study adopts the high flexibility of GPR learning. Meanwhile, data density control and subset selection ensure data quality and GPR inference efficiency.

4) \textit{Flexibility}. The proposed LAMPC retains the flexibility of AMPC for agent configuration. This method can be applied to systems containing any kind or number of black-box agents.

\section*{Acknowledgment}

The authors would like to acknowledge the financial support of Natural Sciences and Engineering Research Council of Canada (NSERC) and financial and technical support of General Motors.

\bibliographystyle{bibtex/IEEEtran}
\bibliography{bibtex/IEEEabrv,Paper/ref}


\begin{IEEEbiography}[{\includegraphics[width=1in,height=1.25in,clip,keepaspectratio]{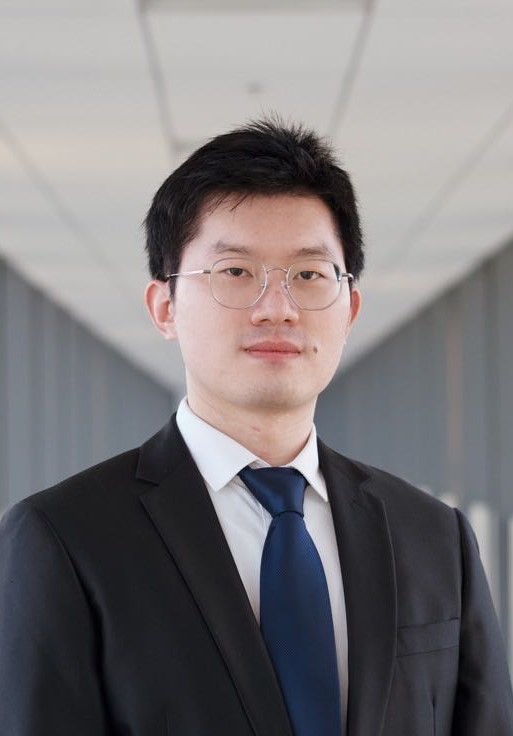}}]{Jiaming Zhong} (Student Member, IEEE) is currently a Ph.D. candidate at the University of Waterloo Mechatronic Vehicle Systems (MVS) Lab. He was also a co-founder and the lead of planning and control of LoopX Innovation Inc. in Ontario, Canada. He received his B.S. and MASc. degrees in mechanical engineering from the Beijing Institute of Technology, China, in 2014 and 2017. He previously worked as a senior software engineer in SAIC Motor Co., Ltd. and NIO Co., Ltd. in Shanghai, China. His research interests include learning-based planning and control, multi-agent theory, and autonomous driving.
\end{IEEEbiography}

\begin{IEEEbiography}[{\includegraphics[width=1in,height=1.25in,clip,keepaspectratio]{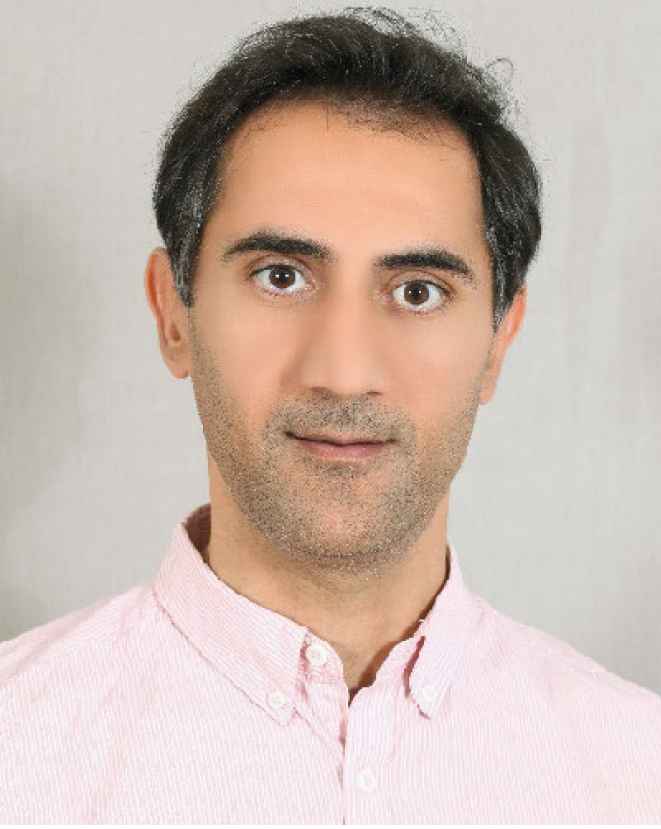}}]{Reza Valiollahi Mehrizi} received the M.Sc. and Ph.D. degrees in statistics from the University of Waterloo, ON, Canada, in 2017 and 2021, respectively.
He is currently employed as a Data Scientist and a Statistical Researcher with the Mechatronic
Vehicular System Laboratory, University of Waterloo. His research interests include building
and developing predictive models using machine learning methods, deep learning object detection and semantic image segmentation, experimental designs, and survival analysis.
\end{IEEEbiography}


\begin{IEEEbiography}[{\includegraphics[width=1in,height=1.25in,clip,keepaspectratio]{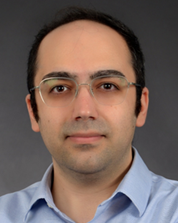}}]{Mohammad Pirani} (Member, IEEE) is an assistant professor with the Department of Mechanical Engineering, University of Ottawa, Canada. He was a research assistant professor in the Department of Mechanical and Mechatronics Engineering at the University of Waterloo (2022–2023). He held postdoctoral researcher positions at the University of Toronto (2019–2021) and KTH Royal Institute of Technology, Sweden (2018–2019). He received a MASc degree in electrical and computer engineering and a Ph.D. degree in Mechanical and Mechatronics Engineering, both from the University of Waterloo in 2014 and 2017, respectively. His research interests include resilient and fault-tolerant control, networked control systems, and multi-agent systems.
\end{IEEEbiography}

\begin{IEEEbiography}[{\includegraphics[width=1in,height=1.25in,clip,keepaspectratio]{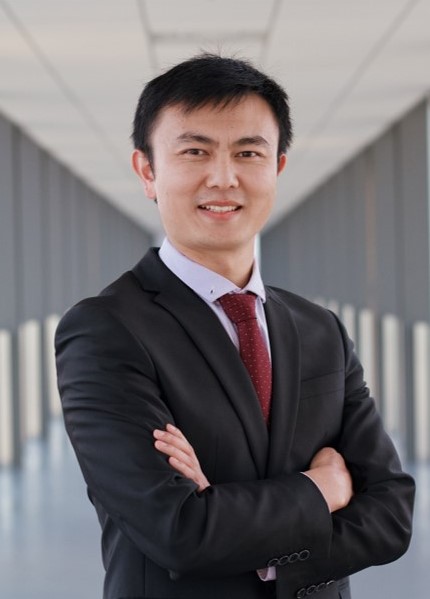}}]{Chao Yu} (Student Member, IEEE) is a PhD candidate in Mechanical and Mechatronics Engineering at the University of Waterloo. He is also the founder and the CEO of LoopX Innovation Inc., an autonomy and digital solution provider in Ontario, Canada. He previously worked in GM Motor China and USA. His research interests focus on machine learning-based control systems for vehicles.
\end{IEEEbiography}

\begin{IEEEbiography}[{\includegraphics[width=1in,height=1.25in,clip,keepaspectratio]{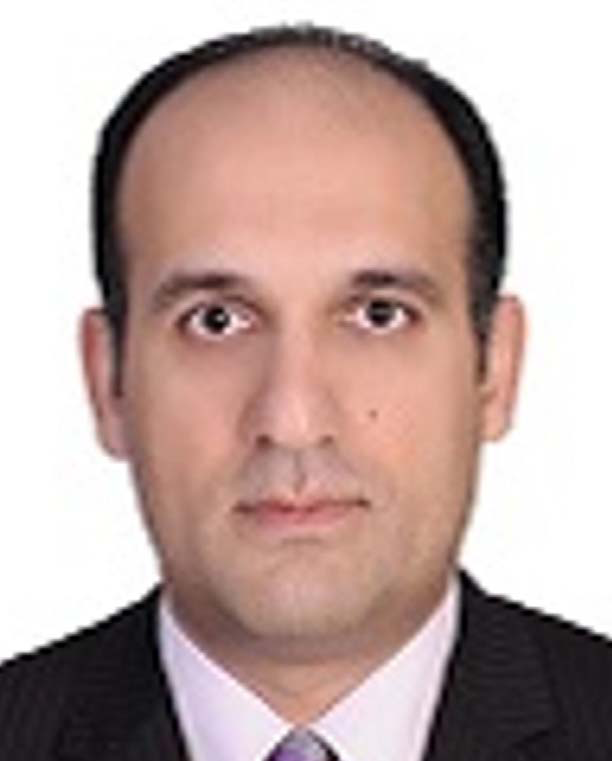}}]{Alireza Kasaiezadeh} received the B.Sc. and M.Sc. degrees in mechanical engineering from the Sharif
University of Technology, Tehran, Iran, in 1999 and 2001, respectively, and the Ph.D. degree in mechanical and mechatronics engineering from the University of Waterloo, Waterloo, ON, Canada,
in 2012. He is currently a Staff Researcher with the GM Technical Center, General Motors Company, Warren, MI, USA, where he is involved in the area of automated driving and vehicle control. His current research interests include automated driving, vehicle dynamics, and control and optimization.
\end{IEEEbiography}

\begin{IEEEbiography}[{\includegraphics[width=1in,height=1.25in,clip,keepaspectratio]{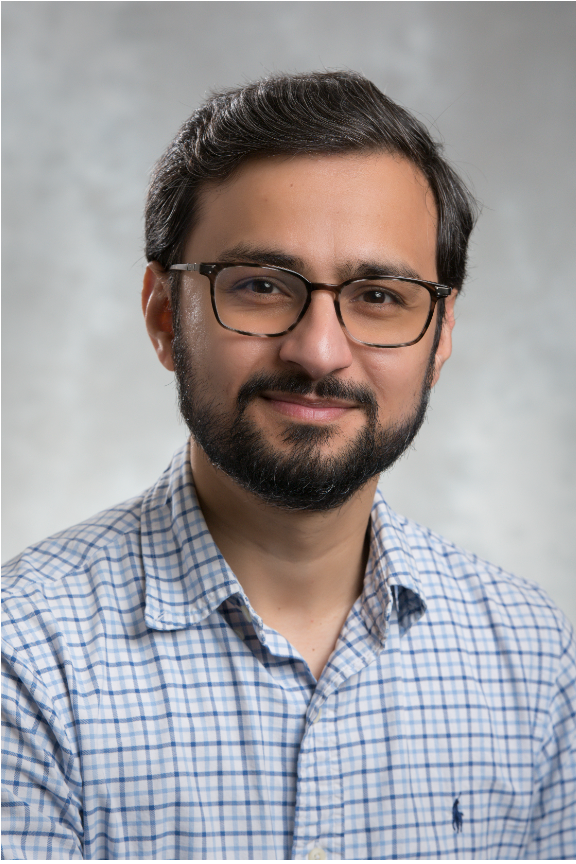}}]{Yash Vardhan Pant} is an Assistant Professor in the Department of Electrical and Computer Engineering at the University of Waterloo. His research focuses on building robust and reliable autonomous systems, using elements of Control Theory, Formal Methods, Machine Learning and Optimization. He received a PhD in Electrical Engineering from the University of Pennsylvania in 2019, where he was a recipient of the Richard K. Dentel memorial award for research in Urban transportation. Prior to joining Waterloo in July 2021, he was a postdoctoral fellow at the Department of Electrical Engineering and Computer Sciences at the University of California, Berkeley.
\end{IEEEbiography}

\begin{IEEEbiography}[{\includegraphics[width=1in,height=1.25in,clip,keepaspectratio]{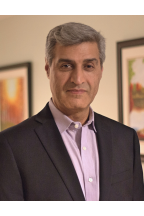}}]{Amir Khajepour} (Senior Member, IEEE) is a professor of Mechanical and Mechatronics Engineering and the Director of the Mechatronic Vehicle Systems (MVS) Lab at the University of Waterloo. He held the Tier 1 Canada Research Chair in Mechatronic Vehicle Systems from 2008 to 2022 and the Senior NSERC/General Motors Industrial Research Chair in Holistic Vehicle Control from 2017 to 2022. His work has led to the training of over 150 PhD and MASc students, filing of 30 patents, publication of 600 research papers, numerous technology transfers, and the establishment of several start-up companies. He has been recognized with the Engineering Medal from Professional Engineering Ontario and is a fellow of the Engineering Institute of Canada, the American Society of Mechanical Engineering, and the Canadian Society of Mechanical Engineering.
\end{IEEEbiography}

\end{document}